%% file: sample-base.tex
\documentclass[sigconf,screen,noacm]{acmart}

\usepackage{booktabs, makecell, tabularx} 

\setcopyright{none}
\renewcommand\footnotetextcopyrightpermission[1]{}
\acmISBN{978-1-4503-8713-2/22/04}

\acmConference[SAC'22]{ACM SAC Conference}{April 25 –April 29, 2022}{Brno, Czech Republic}
\acmYear{2022}
\copyrightyear{2022}

\acmArticle{4}
\acmPrice{15.00}

\begin{document}
\makeatletter
\def\@copyrightspace{\relax}
\makeatother

\title{Attention-based Clinical Note Summarization}
  
\renewcommand{\shorttitle}{SIG Proceedings Paper in LaTeX Format}

\author{Neel Kanwal}
\affiliation{%
  \institution{University of Stavanger}
  \country{Norway}
  \city{Stavanger}
  \postcode{4008}}
\email{neel.kanwal@uis.no}
\orcid{0000-0002-8115-0558}

\author{Giuseppe Rizzo}
\affiliation{%
  \institution{Links Foundation}
  \city{Torino}
  \country{Italy}}
\email{giuseppe.rizzo@linkfoundation.com}

\renewcommand{\shortauthors}{Kanwal and Rizzo}

\begin{abstract}
In recent years, the trend of deploying digital systems in numerous industries has hiked. The health sector has observed an extensive adoption of digital systems and services that generate significant medical records. Electronic health records contain valuable information for prospective and retrospective analysis that is often not entirely exploited because of the complicated dense information storage. 
The crude purpose of condensing health records is to select the information that holds most characteristics of the original documents based on a reported disease. 
These summaries may boost diagnosis and save a doctor's time during a saturated workload situation like the COVID-19 pandemic. 
In this paper, we are applying a multi-head attention-based mechanism to perform extractive summarization of meaningful phrases on clinical notes. Our method finds major sentences for a summary by correlating tokens, segments, and positional embeddings of sentences in a clinical note. The model outputs attention scores that are statistically transformed to extract critical phrases for visualization on the heat-mapping tool and for human use. 
\end{abstract}

%



\maketitle

\pagestyle{plain}

\input{samplebody-conf}

\bibliographystyle{unsrt}
\bibliography{sample_base}

\end{document}

%% file: samplebody-conf.tex
\section{Introduction}
\label{sec:1}

Presenting text in a shorter form has been practiced in human history long before the birth of computers. A \textit{summary} is defined as a document that conveys valuable information with significantly less text than usual~\cite{10.1162/089120102762671927}. Summarization can be sensitive in the medical domain due to medical abbreviations and technicalities. 
The summarization task can be categorized into two categories from a linguistic perspective. \textit{Extractive} summarization is an indicative approach where phrases are scored based on similarity weights and chosen to produce verbatim. Contrarily, \textit{abstractive} summarization is an informative approach that requires understanding a topic and generating a new text using fusion and compression. It relies on novel phrases, lexicon, and parsing for language generation~\cite{article}. 

Natural language processing (NLP) has been valuable to clinicians in saturated work environments. For instance, health information systems have reduced the workload of doctors during the Coronavirus (COVID-19) pandemic. Therefore, a clinical note summarizer can be helpful in a similar fashion. The notion of presenting a condensed version of literature abstracts using computers and algorithms became part of significant research in the late 1950s~\cite{5392672}. This approach was based on finding word frequency and scoring their significance. This idea later evolved in finding summaries based on grammatical position in the text~\cite{5392648}. In contrast, other works~\cite{10.3115/990820.990892, strzalkowski1996natural} proposed query-based summarization frameworks, similar to information retrieval techniques. These methods were similar to multi-query vector in multi-head attention for mapping using a query and key-value pair to an output. \par

Meanwhile, some fundamental questions that arise while summarization are i) which content is essential to select and ii) how to create a shorter version of it ~\cite{inbook}. With rising popularity of transformer~\cite{transformer} as a tool for various text analysis tasks~\cite{bahdanau}. Our work proposes a transformer-based method for selecting meaningful phrases from clinical discharge summaries and extracting them by preserving the sense of clinical notes based on the identified disease.  

Inspired by~\cite{clinicalbert}, we have fine-tuned a Bidirectional Encoder Representation Transformer (BERT) model on discharge notes of the Medical Information Mart for Intensive Care (MIMIC-III) dataset. These discharge notes are classified by diseases and capture important syntactic information based on International Classification of Diseases (ICD-9) labels. We extract a discrete attention distribution from the first head of the last layer. This probabilistic distribution is later translated using power transforms~\cite{book} to create a monotonic attention distribution over bell curve~\cite{attentions}. Finally, the summary comprises sentences with attention scores higher than the mean attention scores of all other sentences in the original clinical note.\par

This paper is organized as follows: Section~\ref{sec:2} illustrates various extractive summarization approaches and their implementations on medical documents. Section~\ref{sec:3} describes the methodology for extractive summarization task. Section~\ref{sec:4} presents various evaluation methods and states a suitable method for this task. We have displayed results in section~\ref{sec:5}.
Finally, we conclude the discussion in section~\ref{sec:6} with limitations possible future directions in section~\ref{sec:7}.\par


\section{Related Work} 
\label{sec:2}
The early works on summarization are based on many different surface-level approaches for the intermediate representation of text documents. These methods focus on selecting top sentences based on greedy algorithms and aim to maximize coherence and minimize redundancy~\cite{Allahyari, zhong2020extractive}. These techniques can be further generalized into:
\begin{itemize}
    \item \textbf{Corpus-based Approach:} It is a frequency-driven approach based on common words often repeated and do not carry salient information. It relies on an information retrieval paradigm in which common words are considered query words. SumBasic~\cite{Vanderwende} is a similar centroid-based method that uses word probability as a factor of importance for sentences. Words in each centroid with higher probabilistic values are selected for a summary. \par
    
    \item \textbf{Cohesion-based Approach:} Some techniques fail when extraction is bound to anaphoric~\footnote{Anaphoric expressions are words that relate the sentences using pronouns such as he, himself, that.} expressions or lexical chains to relate two sentences. Brin et al.~\cite{Brin} proposed a co-reference system that uses cohesion in web search. In clinical notes, anaphoric expressions are frequently used, but they refer to the same subjects meaning that we have a relation to the patient only. \par
    
    \item \textbf{Rhetoric-based Approach:} This approach relies on forming text organization in a tree-like representation~\cite{Goldstein,  Kupiec}. Text units are extracted based on their position close to the nucleus. For clinical summaries, we often have multiple nuclei for different diseases.\par
    
    \item \textbf{Graph theoretic Approach:} A few popular algorithms like HITS~\cite{Knight} and Google`s PageRank~\cite{mihalcea} instigated base for graph-based summarization. It helps to visualize intra-topic similarity where nodes present several topics, and vertices show their cosine similarity with sentences~\cite{gupta}. It makes visual representation easy with different tools like MDIGESTS~\cite{Chongtay}.\par
    
    \item \textbf{Machine Learning (ML) Approach:} ML models outperform nearly all kinds of tasks, including text summarization. Neural networks (NN) can better exploit hidden features from the text. Attention mechanism coupled with convolution layers helps to choose important phrases based on their position in the document. A recent trend of analyzing text using Bayesian models has also gained popularity~\cite{Ayodele}. Miller et al.~\cite{miller2019leveraging} used BERT to make text encoding and applied K-means to find sentences from health informatics lectures. Their model offered a weakness for large documents since the extraction ratio is fixed for K sentences. Liu et al.~\cite{liu2019fine} trained an extractive BERT model from abstractive summaries using a greedy method to generate an oracle summary for maximizing the ROGUE score. BERTSUM~\cite{liu2019fine} used a trigram blocking method to extract candidate sentences based on golden abstractive summaries in CNN/daily mail dataset. 
\end{itemize}

\subsection{Implementations for Medical Documents}
Clinicians heavily rely on text information to analyze the condition of the patient. Vleck et al.~\cite{Vleck} followed a cognitive walk-through methodology by identifying specific relevant phrases to medical understanding. Laxmisan et al.~\cite{Laxmisan} formed a clinical summary screen to integrate with health systems. The core purpose was to avail more interaction time for a clinician. \par

Feblowitz et al.~\cite{Feblowitz} proposed a five-stage architecture to facilitate clinical summarization tasks. Their framework (AORTIS) described distinct phases, namely Aggregation, Organization, Reduction, Interpretation and Synthesis (AORTIS)~\cite{wright}. It was based on producing short laboratory reports. The AORTIS model was assessed and validated using Cohen`s Kappa index. Alsentzer et al.~\cite{Alsentzer} did a similar job using Bayesian modeling. Pivovarov et al.~\cite{Pivovarov} employed heterogeneous sampling and topic modeling. Their approach materialized a Concept Unique Identifier (CUI) upper bound to choose a phrase with a high probability of being classified as a disease.\par

Thomas et al.~\cite{Thomas} proposed a semi-supervised graph-based method for summarization using NN and node classification. The model was limited to datasets other than the clinical domain. Other researchers followed a similar kind of approach for Multi-document summarization~\cite{Christensen, Michihiro}. Azadani et al.~\cite{Azadani} later carried out these ideas to biomedical summarization. Their model uses graph clustering that forms a minimum spanning tree using Unified Medical Language System (UMLS).\par

Another ontology-oriented graphical representation method uses clustering to form data centrality and mutual refinement~\cite{Yoo}. Mis-classification index (MI) was used as a primary evaluation metric to verify cluster purity. Generally, graphical methods have concluded better results in most of the cases in the literature.

\begin{figure*}[ht]
    \centering
    \includegraphics[width =0.99 \textwidth,height = 12cm ]{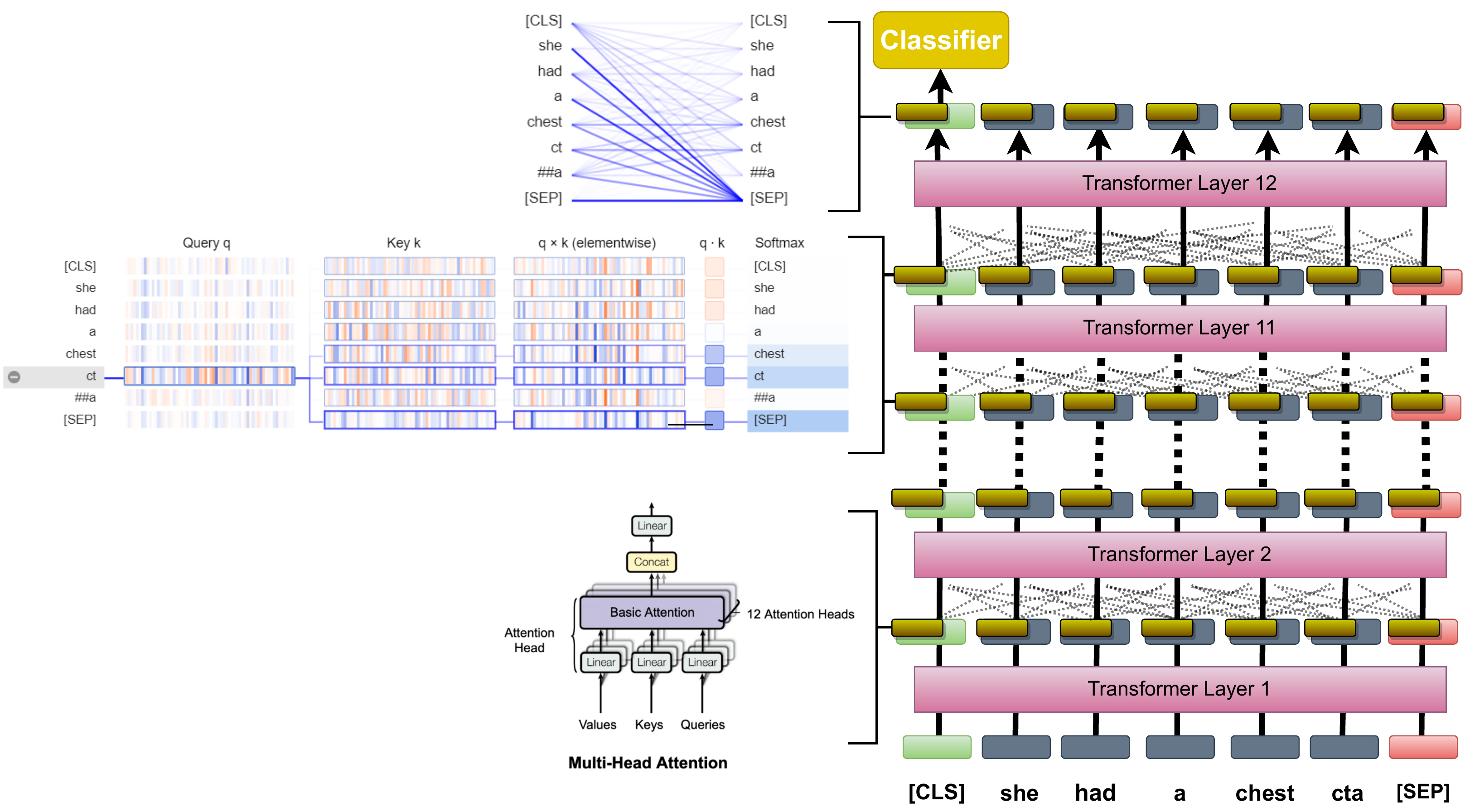}
    \caption{BERT base architecture with 12 transformer layers. Every layer carries embeddings (blue) and attention scores (green) for corresponding tokens. This multi-head attention correlates to every other word as seen in the images at left. The classifier token from the last layer is used for every sentence as detailed in the methodology.}
    \label{fig:1}
\end{figure*}

\section{Methodology} \label{sec:3}
Our approach uses a base BERT-model fined-tuned on ICD-9 labeled MIMIC-III discharge notes. The model was trained mainly to identify ICD-9 labels based on described symptoms and diagnostic information. The model outputs attention scores for all sentences from discharge notes. We extract sentences whose attention scores are higher than the mean value of all other sentences in the original note. The model is compared against three baselines ~\cite{Edmundson, Michihiro, miller2019leveraging} using divergence methods of word probability distributions for quantitative analysis. Our summarization approach works effectively when reference or human-made summaries are not available. Furthermore, table~\ref{results2} represents qualitative analysis against chosen baseline approaches.\par

\subsection{Dataset} \label{sec:3.1}
MIMIC-III is an open-access publicly available database of unstructured and unidentified health records~\cite{Johnson}. It is an extensive relational database with 26 tables linked with subject identity. It includes raw notes of 36,998 patients for each hospital stay in the "EVENTS" table (see~\footnote{https://physionet.org/content/mimiciii/1.4/}). Each discharge note is tagged with a unique label for the identified disease. In total, we have 47,724 clinical discharge notes that comprise several details from radiology, nursing, and prescription. These notes can be equated with a multi-topic document based on multiple labels for each medical note. Moreover, MIMIC-III was only published for labeling diagnostics and does not contain reference summaries. \par

\subsection{Neural Architecture} \label{sec:3.2}
We have utilized a neural architecture that is built on top of transformer~\cite{transformer}. BERT is multi-layer neural architecture and has two major variants. We have used a base variant with 12 layers (transformer blocks), 768 hidden sizes, 12 attention heads, and 110 million parameters. The language model has shown significant improvements in various language processing tasks with fine-tuning~\cite{Howard, Dai}. 
The BERT model usually creates embeddings in both directions for the representation of the inputs~\cite{DBLP:journals/corr/abs-1810-04805}. We have found that attention heads corresponding to delimiter tokens are remarkably effective for semantic understanding. \par

\subsection{Implementation Details}\label{sec:3.3}
The work presented in this paper employs a fine-grained understanding of the notes to signify sentences relevant to the classification and brings more information to a summary. A reduced sample of randomly selected 100 notes from the MIMIC-III dataset is selected to quantitatively and qualitatively assess the model's performance. Finally, we have demonstrated attention scores for sentences using a highlighting tool (see sec.~\ref{sec:3.4}) to inspect the output results. Figure~\ref{fig:1} shows the model along with attention flow.\par

\paragraph{\textbf{Pre-processing: }} Clinical documents contain many irregular abbreviations and periods for their particular formatting. Some notes are in grammatical order, whereas other parts are written as review keywords. We have used the custom tokenizer presented in~\cite{Mullenbach} to formulate data as lists of sentences. It removes tokens with no alphabetic character or a percentage of drug prescriptions. \par

\paragraph{\textbf{Token Representation: }} A sentence flows downstream as a sequence of tokens accompanied by two additional unique tokens. An input representation for any token is formed by combining token, position, and segment embedding. [CLS] is the first token that classifies a sentence and appears initially. [SEP] is a separator token used to identify the end of the stream. The output [CLS] representation can be fed to the classifier for different tasks.\par

\paragraph{\textbf{Fine-Tunning: }} Fine-tuning has a huge effect on performance for supervised tasks~\cite{kovaleva}. We have fine-tuned the BERT model on the entire MIMIC-III using maximum sequence length, batch size 8, learning rate 3e-5, ADAM optimizer with epsilon 1e-8, and keeping other hyper-parameters the same as that of pre-training. The fine-tuned model can classify [CLS] token for maximum-likelihood of ICD-9 label. \par

\paragraph{\textbf{Attention Extraction: }} Fine-tuning helps to encode semantic knowledge in self-attention patterns~\cite{kovaleva}. The multi-head attention mechanism embeds an attention score in tokens of every sentence in the clinical note. Since the last layer of the BERT model is considered vital to the task~\cite{van2019does}, we capture the first attention head of the last layer for cross-sentence relation as observed with BertViz~\cite{Vig}. This attention head focuses on a special [CLS] token corresponding to the whole sentence. The obtained attention score for a sentence is used as a measure of significance in a clinical note. Equation~\ref{eq:1} and Equation~\ref{eq:2} show how the dot-product attention is calculated in layers.\par

 \begin{equation}\label{eq:1}
      a_i = Softmax(f(Q,K_i)) = \frac{exp(f(Q,K_i))}{\sum _i exp(f(Q,K_i))}
 \end{equation}
 
 \begin{equation}\label{eq:2}
      Attention(Q,K,V) = \sum _i a_i\ast V_i
 \end{equation}
 
\begin{figure}[t!]
    \centering
    \includegraphics[width=8cm,height=6cm]{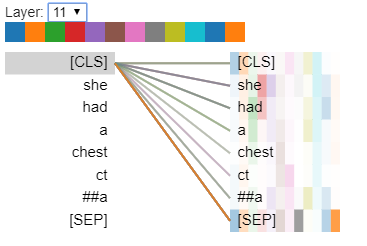}
    \caption{Attention visualization shows how every attention-head in BERT architecture finds words useful compared to other words in a sentence. We have a different color for every head, identifying its position in the last layer. We have applied the idea to choose useful sentences compared to other sentences in the original note. The demonstration is performed using BertViz tool~\cite{Vig}.}
    \label{fig:2}
\end{figure}

 In a nutshell, a set of pre-processed clinical notes in a list of sentences is fed to the BERT encoder, creating embedding and attention scores at each layer. The attention score corresponding to [CLS] decides whether a sentence is a good candidate for the summary. Figure~\ref{fig:2} illustrates the positional relevance of the [CLS] token. We have later selected sentences whose attention scores are above the average attention value of other sentences in the original note. This extraction incentives dynamic selection, unlike a fixed sentence summary ratio. For example, a sentence with an attention score of 0.14 is chosen if it is greater than the average of attentions of all sentences in the document.\par 
 
\subsection{Sentence Attention Visualization}\label{sec:3.4}
The attention distribution for tokens on the last layer has an irregular pattern. In order to perform a heat-mapping, we have utilized a tool, namely Neat-Vision~\footnote{\url{https://github.com/cbaziotis/neat-vision}}.

\begin{figure}[t]
    \centering
    \includegraphics[width=8cm,height=6cm]{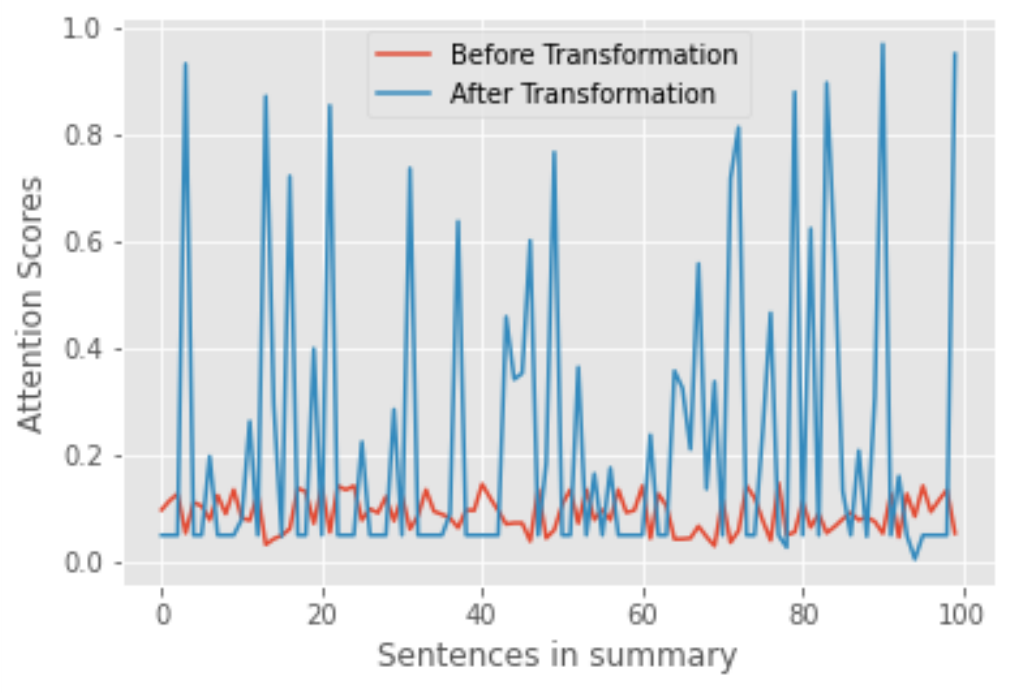}
    \caption{Effect of quantile transform on attention scores of all sentences in the document. The red line shows obtained attention distribution with low variance, whereas the blue line shows higher variance that can be efficiently utilized for heat-mapping.}
    \label{fig:3}
\end{figure}

\begin{figure}[t]
 \centering
 \includegraphics[width=8cm]{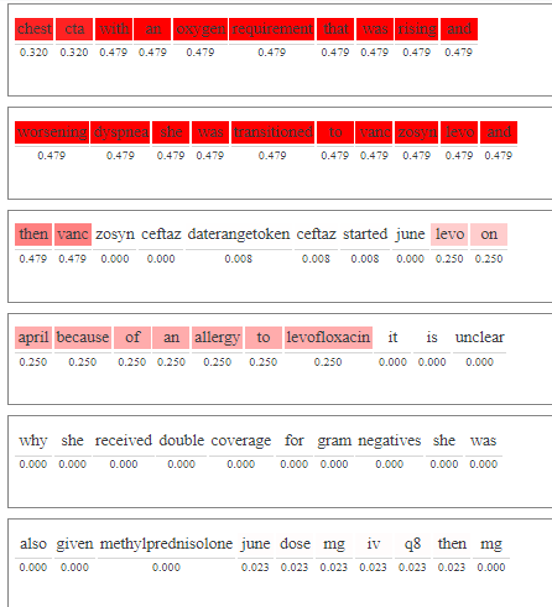}
 \caption{An excerpt of heat-mapping on Neat-Vision tool with transformed attention values to highlight the importance of sentence with red color.}
 \label{fig:4}
\end{figure}

This tool requires a fixed input format and outputs a text heat-map. It demands that input data be organized in a particular structure for vibrant coloring. We have stratified the distribution obtained from the neural architecture to the Gaussian distribution using the \textit{Quantile Transformation}~\cite{book}. This turns a sentence with great attention more fragrant in visibility and vice versa. In other words, it makes sentences with higher attention scores rosier than the other ones in the Neat-Vision tool. Figure~\ref{fig:3} shows the impact of transformation on attention series data. Here x-axis presents the number of sentences, and the y-axis accounts for the value of the attention score for the corresponding sentence. This demonstration will considerably impact clinician practice by alienating time spent while reading long health records. Figure ~\ref{fig:4} exhibits the usefulness of heat-mapping concepts for health systems.\par

\section{Evaluation} \label{sec:4}
Evaluation in summarization has been a critical issue, mainly due to the absence of a gold standard. Many competitions such as DUC~\footnote{\url{http://duc.nist.gov/}}, TREC~\footnote{ \url{http://trec.nist.gov/}}, SUMMAC~\footnote{\url{https://www-nlpir.nist.gov/related_projects/tipster_summac/}} and MUC~\footnote{\url{http://www.itl.nist.gov/iad/894.02/related projects/muc/proceedings/muc 7 toc.html}} propose different metrics. The interpretation of these metrics is not very simple, mainly because the same summary receives different scores under different measures. Automatic evaluation for the quality of the summary is an ambitious task and can be performed by making a comparison with a human-generated summary. For this reason, evaluation is normally limited to domain-specific and opinion-oriented areas~\cite{inbook}.\par

Formally, these evaluation methods can be divided into two areas. In \textit{extrinsic} evaluation, summaries are manually analyzed for the original document. For instance, a clinician can do this in our case and may result in different opinions based on his understanding. Miller et al.~\cite{miller2019leveraging} leveraged this manual clinical evaluation to compare the performance of their model. In \textit{intrinsic} evaluation, the extracted summary is directly matched with the ideal summary created by humans. The latter can be divided into two classes, primarily because it is hard to establish an ideal reference summary by a human. 
\begin{itemize}
\item \textit{Text quality Evaluation:} It is more related to linguistic check that examines grammatical and referential clarity. This assessment is not comprehensive since medical summaries are unstructured documents with many abbreviations and clinical jargon.
\item \textit{Content-based Evaluation:} It rates summaries based on provided reference summaries~\cite{steinberger}. Some of the common approaches are ROGUE (Recall-Oriented Understudy for Gisting Evaluation)~\cite{Lin}, Cosine Similarity and Pyramid Method~\cite{nenkova2007pyramid}. Liu et al.~\cite{liu2019fine} studied unigram and bigram ROGUE overlaps for different components of BERTSUM for their single-document summaries.
\end{itemize}

Sripada et al.~\cite{sripada} in their work exemplified that a summary can be considered adequate if it has a similar probability distribution as that of the original document. The hypothesis was compared in other works~\cite{nenkova, yih} where this light-weight and less complex method demonstrated more refined results. This criterion is a more practical evaluation for our methodology since we do not have the reference summaries. We will compare the distribution of words in the original and summary document to identify their effectiveness. We will use two tests for evaluating the goodness of our synopsis, namely Kullback–Leibler divergence (KLD)~\cite{KLD} and Jensen–Shannon divergence (JSD)~\cite{JSD}.\par

\paragraph{KL Divergence:} It is a measure of the difference between two distributions. This measure is asymmetric, and the minimum KLD value shows better relative interference for distributions; for discrete probability distributions P and Q mapped on probability space $\zeta$, KLD from Q to P is defined in Equation ~\ref{eq:3} ~\cite{das}; \par

\begin{equation}\label{eq:3}
\centering
\Large{KLD(P || Q) = \sum_{x\epsilon\zeta}P(x)\log{\frac{P(x)}{Q(x)}}}  
\end{equation}

\paragraph{JS Divergence:} It is an extension of KL divergence that quantifies the difference in a slightly modified way. It is a smoothed and normalized form, assuring symmetry among inputs as reported in Equation ~\ref{eq:4}~\cite{JSD}.\par

\begin{equation}\label{eq:4}
\centering
\large
{JSD(P || Q) = \frac{1}{2}KLD(P || M) + \frac{1}{2}KLD(Q || M).}  
\end{equation}
where, \par

\begin{equation}\label{eq:5}
\centering
 M = \frac{1}{2} (P + Q)
\end{equation}


\section{Results} \label{sec:5}
We have presented results for comparing our model with three baseline extractive summarization methods. First, Part-of-Speech-based sentence tagging is established on the empirical frequency selection method. The second one centers around the graphical method, and the third one uses BERT combined with K-means to find top k sentences in a centriod. The results can be reproduced, and source code for replication is available on Github~\footnote{\url{https://github.com/NeelKanwal/BERTOLOGY-Based-Extractive-Summarization-for-Clinical-Notes}} repository.\par

The proposed architecture shows significant improvement compared to baseline approaches. Divergence scores show how estimating differences in distributions can help in anticipating the word distribution of both documents. Table~\ref{results1} exhibits that our extracted summaries are more informative than others based on a lower average of KL and JS divergence scores. Frequency-based approach outcomes highest divergence among others. JSD and KLD scores for the graph-based method show a relative amelioration compared to frequency-based methods. There is a bit of quantitative difference in values with the centriod-based K-means approach due to their nature of calculating sentence embeddings in a similar way. Our method is dynamic in choosing the length of the summary, which overcomes the weakness of fixed K sentences described in the paper~\cite{miller2019leveraging}. Overall, the attention mechanism poses great abstraction power for the summarization task.\par

\begin{figure}[t]
 \centering
 \includegraphics[width=8.5cm,height=6.5cm]{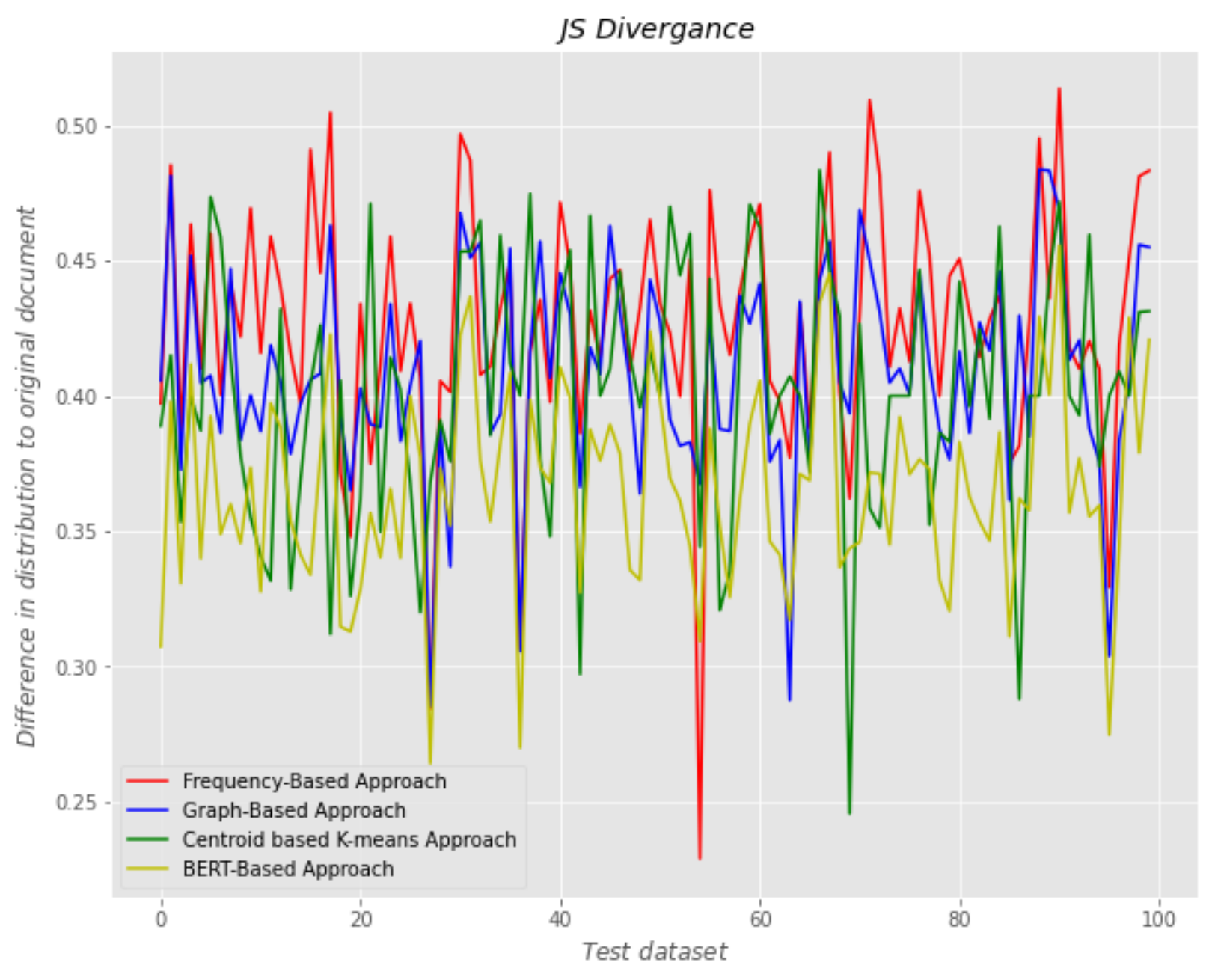}
 \caption{Line chart for JSD values for experimented models over sampled-set of clinical notes from MIMIC-III dataset.}
 \label{fig:5}
\end{figure}

\begin{figure}[t]
 \centering
 \includegraphics[width=8.5cm,height=6.5cm]{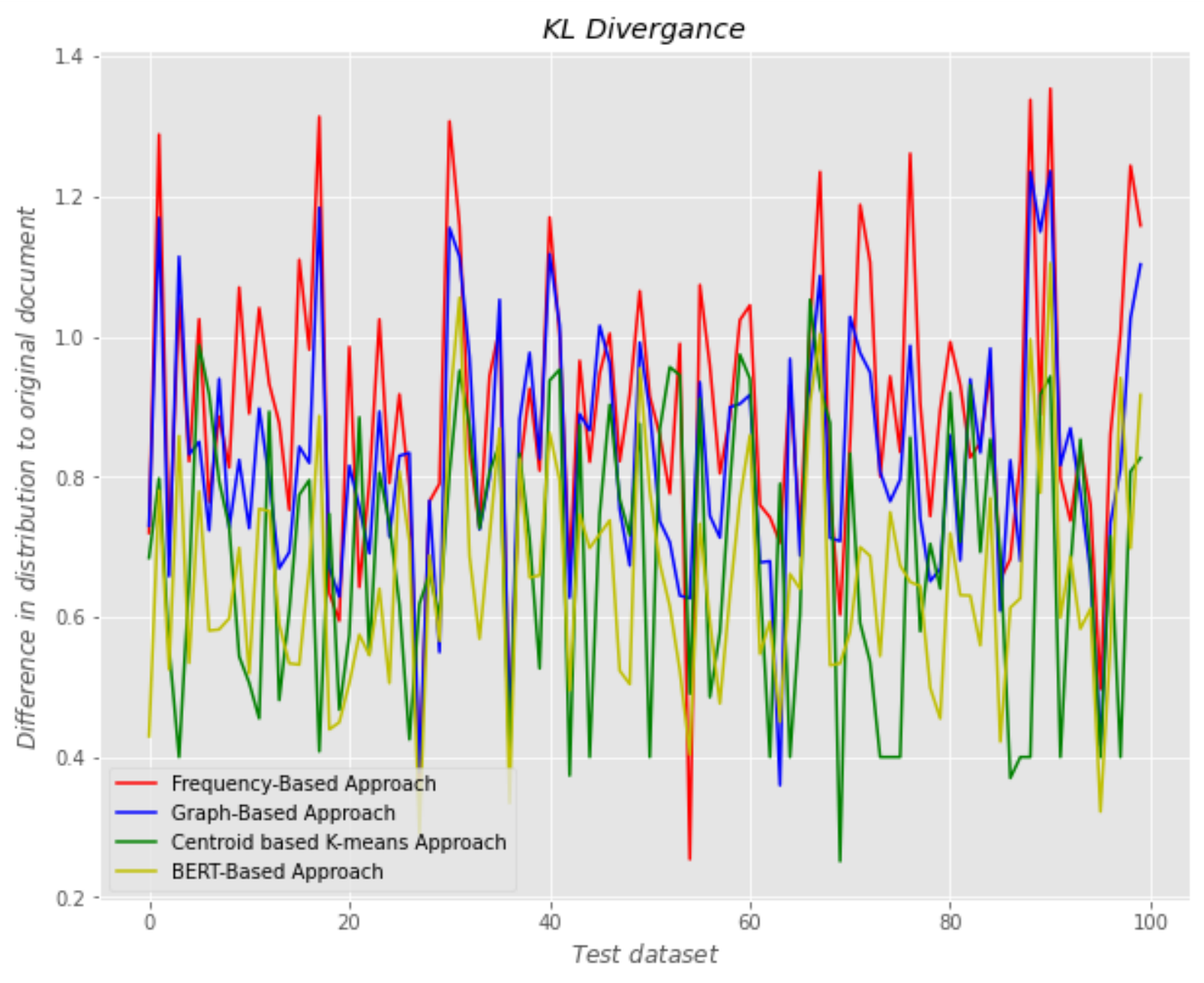}
 \caption{Line chart for KLD values for experimented models over sampled-set of clinical notes from MIMIC-III dataset.}
 \label{fig:6}
\end{figure}

\begin{table}[h!]
    \large\centering
 \begin{tabular}{|l| c c |}
    \hline
    Models & \textbf{KLD$\downarrow$} & \textbf{JSD$\downarrow$}\\
    \hline \hline
    Frequency-Based Approach & 0.892 & 0.426  \\
    Graph-Based Approach & 0.827 & 0.408  \\
    Centroid-Based K-means Approach & 0.80 & 0.41 \\
    \textbf{Our Proposed Architecture} & \textbf{0.795} & \textbf{0.405} \\[0.5ex]
    \hline
    \end{tabular}
    \caption{Experimental results on a reduced sample-set of 100 random clinical notes from MIMIC-III dataset compared with Frequency-Based Approach ~\cite{Edmundson}, Graph-Based Approach ~\cite{Michihiro} and Centroid based K-means Approach~\cite{miller2019leveraging} using KLD and JSD Values. A lower value pertains to a better-correlated summary.}
    \label{results1}
\end{table}

\begin{table*}[t]

\begin{tabular}{|| p {17cm} ||}
\hline\hline
\vspace{0.5em}\textbf{Centroid-Based K-means Summary: }daily disp tablet delayed release e.c. lastnametoken on february at 15pm cardiologist dr. lastnametoken on february at 30am wound check on thurs january at am with cardiac surgery on hospitaltoken please call to schedule appointments with your primary care dr. lastnametoken in march weeks please call cardiac surgery office with any questions or concerns telephonenumbertoken answering service will contact on call person during off hours completed by january.\newline\\

\textbf{Frequency-Based Summary: }disp tablet refills ranitidine hcl mg tablet sig one tablet daily. refills tramadol tablet two tablet q6h hours as needed for pain. tablet senna mg tablet  One tablet daily, disp tablet refills furosemide mg tablet for Mitral valve repair coronary artery bypass. graft x left internal mammary artery to left anterior descending history of present illness year old female who was told she had mvp since age currently quite active but has noticed some dyspnea on exertion when walking up hills most recent echo revealed severe mvp and moderate to severe Daily daily disp tablet delayed release e.c. s refills docusate sodium mg capsule sig one capsule po bid times Disp tablet er particles crystals s refills discharge disposition home with service facility hospitaltoken vna discharge diagnosis mitral regurgitation coronary artery  disease. \newline\\

\textbf{Graph-Based Summary: }refills docusate sodium mg capsule one capsule a day magnesium hydroxide suspension thirty ml at bedtime as needed for constipation atorvastatin tablet one tablet daily. disp tablet s refills furosemide tablet once a day for days disp tablet refills ranitidine hcl mg tablet daily.
please shower daily including washing incisions gently with mild soap no baths or swimming until cleared by surgeon look at your incisions daily for redness or drainage. please no lotions cream powder or ointments to incisions each morning you should weigh yourself and then in the evening take your temperature these should be written down on the chart no driving for approximately one month and while taking narcotics will be discussed at follow up appointment with surgeon when you will be able to drive no lifting more than pounds for weeks please call with any questions or concerns telephonenumbertoken females please wear bra to reduce pulling on incision avoid rubbing on lower edge. please call cardiac surgery office with any questions or concerns telephonenumbertoken answering service will contact on call person during off hours followup instructions you are scheduled for the following appointments surgeon dr. lastnametoken on february at 15pm cardiologist dr. lastnametoken on february at 30am wound check on thurs january at am with cardiac surgery on hospitaltoken.\newline \\

\textbf{Our Proposed Approach: }old female who was told she had mvp currently quite active but has noticed some dyspnea on exertion when walking up hills. she presents for surgical consultation past medical history mitral regurgitation copd secondary to asbestos exposure as a child arhtritis cataracts headaches lactose intolerance r wrist and elbow surgery. widowed occupation retired disabled nurse tobacco quit smoking in father died suddenly at cause unknown physical exam. no spontaneous echo contrast is seen in the left atrial appendage there is a small pfo with left to right flow overall left ventricular systolic function is normal lvef in the face of mr there is normal free wall contractility there are simple atheroma in the descending thoracic aorta the aortic valve leaflets are mildly thickened trace aortic regurgitation is seen the posterior leaflet is very degenerate and there is moderate to severe mitral regurgitation. there is no pericardial effusion the tip of the sgc is seen at the pa bifurcation post cpb the patient is av paced on no inotropes the pfo is closed normal biventricular systolic fxn there is a mitral ring prosthesis which is well seated trace mr residual mean gradient with an area of no ai aorta intact. mrs. lastnametoken was a same day admit after undergoing all pre operative work. she was tolerating a full oral diet her incisions were healing well and she was ambulating in the halls without difficulty it was felt that she was safe for discharge home at this time with vna services all appopriate follow up appointments were arranged.\newline \\

\hline \hline
\end{tabular}
\caption{Qualitative evaluation of our approach with three baseline methods.}
\label{results2}
\end{table*}

As noted in Figure~\ref{fig:5} and ~\ref{fig:6}, there are some summaries where distributional similarity does not outperform due to shorter length. The curve presents that attention-based extraction is more impactful than other counterparts. JS divergence metrics show less fluctuation than other metrics because of their averaging symmetry mechanism. 
Summaries from each method are placed in Table~\ref{results2} for qualitative analysis. It can be observed that summaries generated by our proposed architecture have more coherence and make it easier to adapt clinical understanding. On the other hand, baselines approaches provide short and incoherent sentences for the selected note. Shorter summaries are more likely to lose discriminatory information and affect the degree of understanding; thus, evaluating the usefulness of a summary in terms of sentences may not be optimal. It may be hard for a non-specialist to understand the relative usefulness of each summary as described in section~\ref{sec:4}. This method shows the applicative benefits of dynamic summarization in healthcare systems. Furthermore, it is more helpful for a physician to grab the essence of diagnosis via highlighting tools as displayed in figure~\ref{fig:4}. \par

\section{Conclusion} \label{sec:6}
The immense increase in digital text information has undoubtedly emphasized the need for universal summarization frameworks. Abstractive summarization has been an area of research debatable for specific scenarios, e.g., medical, because of the risk of generating summaries that deliver different meanings of the original notes reported by physicians. However, extractive summarization techniques are relatively reasonable in the clinical domain. At the same time, evaluation in medical summarization is at the most challenging degree compared to other domains. \par

In this paper, we have elucidated a neural architecture for extracting summaries based on multi-head attentions. We have utilized statistical analysis methods to understand the magnitude of relevance between summary and original clinical notes. Our architecture achieves better results on a set of MIMIC-III clinical notes, outperforming frequency, graph-oriented, and centroid-based approaches. The proposed model is domain-specific and outperforms other methods debated in the literature. The evaluation criteria of finding divergence among distributions are suitable when ideal summaries are not present. Furthermore, our proposed model can be integrated into a decision-support system to better interpret clinical information by highlighting diagnostically related phrases.\par


\section{Limitations and Future Work} \label{sec:7}
Medical summarization is a unique and delicate task. It is pretty hard to automatically evaluate whether the obtained summary is a well-condensed representation of the original document. Moreover, performing a qualitative evaluation is labor-intensive and subjective and may also depend on the physician's personal experience with similar diseases. A universal medical summarizer may omit limitations arising from the diverse writing style.\par

The attention-based model is fine-tuned on the MIMIC-III dataset. Therefore, it may not perform well on different clinical notes, written in a different structure and mapped onto a different set of diseases. ICD-9 offers broad coverage and accurate cataloging of diseases; however, we consider ICD-10 in current research activities as more contemporized. A concoction of abstractive and extractive summarization using neural network language generative models may be more bankable for future work.\\
\par